\documentclass[aip,cha,reprint,groupedaddress]{revtex4-1}
\usepackage{float}
\usepackage{epsfig}
\usepackage{natbib}
\usepackage{amsmath}
\usepackage{times}
\usepackage{subfigure}
\usepackage{graphicx}
\usepackage{color}
\usepackage{dsfont}
\usepackage[utf8]{inputenc} 
\usepackage{verbatim}
\usepackage{xcolor}
\usepackage{array, multirow}
\usepackage{booktabs}
\usepackage{bbm}
\usepackage{hyperref}
\usepackage{nicematrix}

\draft 

\begin{document}

\title{Deep Learning-based Analysis of Basins of Attraction}

\author{David Valle}
\email[]{david.valle@urjc.es}
\affiliation{Nonlinear Dynamics, Chaos and Complex Systems Group, Departamento de
F\'{i}sica, Universidad Rey Juan Carlos, Tulip\'{a}n s/n, 28933 M\'{o}stoles, Madrid, Spain\\}

\author{Alexandre Wagemakers}
\email[]{alexandre.wagemakers@urjc.es}
\affiliation{Nonlinear Dynamics, Chaos and Complex Systems Group, Departamento de
F\'{i}sica, Universidad Rey Juan Carlos, Tulip\'{a}n s/n, 28933 M\'{o}stoles, Madrid, Spain\\}

\author{Miguel A.F. Sanju\'{a}n}
\email[]{miguel.sanjuan@urjc.es}
\affiliation{Nonlinear Dynamics, Chaos and Complex Systems Group, Departamento de
F\'{i}sica, Universidad Rey Juan Carlos, Tulip\'{a}n s/n, 28933 M\'{o}stoles, Madrid, Spain\\}

\date{\today}

\begin{abstract}
This research addresses the challenge of characterizing the complexity and unpredictability of basins within various dynamical systems. The main focus is on demonstrating the efficiency of convolutional neural networks (CNNs) in this field. Conventional methods become computationally demanding when analyzing multiple basins of attraction across different parameters of dynamical systems. Our research presents an innovative approach that employs CNN architectures for this purpose, showcasing their superior performance in comparison to conventional methods. We conduct a comparative analysis of various CNN models, highlighting the effectiveness of our proposed characterization method while acknowledging the validity of prior approaches. The findings not only showcase the potential of CNNs but also emphasize their significance in advancing the exploration of diverse behaviors within dynamical systems.
\end{abstract}


\maketitle 

\begin{quotation}
The application of machine learning algorithms to complex dynamical systems has opened up new possibilities for the study and the computation of dynamical systems. One important aspect of studying dynamical systems is understanding their basins of attraction which are the regions in the phase space where initial conditions lead towards a particular attractor. Since we are interested in the asymptotic behavior of trajectories, the basins of attraction provide a visual and analytical tool to have a clear image of which set of initial conditions bring their associated trajectories to specific attractors in phase space.  Basins of attraction can be characterized using various metrics, such as the fractal dimension, the basin entropy, the boundary basin entropy, and the Wada property. These measurements provide a quantitative understanding of the behavior of dynamical systems. Moreover, their prediction involves complex nonlinear operations that are well adapted for machine learning algorithms. Additionally, convolutional neural networks are particularly suitable for this task due to the representation of basins as images, where each pixel corresponds to an initial condition and each color represents a different attractor. The use of machine learning algorithms in this context constitutes an important step towards predicting the sensitivity of the system to initial conditions, and improving our understanding of the behavior of complex dynamical systems.
\end{quotation}

\section{Introduction} \label{sec:Introduction}

Complex dynamical systems are typically modeled through differential or difference equations and therefore follow deterministic rules. However, even though they are deterministic, it does not necessarily mean that these systems are easy to predict. As is well known, in the case of chaotic dynamics, a small variation in the initial conditions may cause that trajectories would go to a different attractor after a sufficiently long run. In particular, in an experimental scenario a slight variation in the initial conditions can completely alter the long-term evolution when the system happens to be chaotic. This inherent property of chaos is closely related to fractal geometry, so it is not surprising that both disciplines are so closely connected~\cite{peitgen1992chaos}.

Fractal geometry and basins of attraction are related concepts in the study of dynamical systems. Fractal geometry is a mathematical framework for describing objects that display self-similar patterns at different scales. In the context of dynamical systems, fractal sets such as the Mandelbrot set are often used to illustrate the behavior of iterated functions~\cite{peitgen1992chaos}. Basins of attraction refer to the regions in the phase space of a dynamical system where initial conditions lead towards a particular attractor, and boundary sets refer to the separation between the different basins~\cite{nusse1996basins, aguirre2009fractal}. The relationship between fractal geometry and basins of attraction lies in the fact that the boundary sets can be fractal, providing a manifestation of the complex and intricate structure of the phase space. In some multistable dynamical systems, the basin boundaries display complicated fractal structures at every scale, hindering the long-term prediction of initial conditions starting in the vicinity of such boundaries.

The analysis of basins of attraction with different metrics offers practical and theoretical information of the dynamical system. The asymptotic behavior of the system is embodied into a single structure that matches an initial condition to an attractor. By tracking the properties of this structure as a system parameter varies, we gain insight into how the system evolves with that parameter modification. This process enriches the understanding of the predictability and stability of the system facilitating the identification of bifurcation points~\cite{Wagemakers2023}.

The unpredictability of a basin can be characterized using various metrics such as the fractal dimension~\cite{mcdonald1985fractal}, the basin entropy~\cite{Daza2016}, the boundary basin entropy, and the Wada property~\cite{Wagemakers2020}. Each measure focuses on different aspects of the unpredictability of the dynamical system in its phase space. The fractal dimension quantifies the long term uncertainty of the initial conditions near a basin boundary. The basin entropy and the boundary basin entropy blend the uncertainty of the boundary, the number of attractors, and topological traits into a single scalar measure. Finally, the Wada property is a topological property associated to dynamical systems with three or more basins, where a boundary point belongs at the same time to all the basin boundaries. The existence of this property implies a higher degree of unpredictability for the dynamical system, as small perturbations near the basin boundaries can lead to any of the different attractors~\cite{Daza2018, DazaClasifying2022, Wagemakers2020}.

Analyzing and predicting the behavior of dynamical systems through basins of attraction presents computational challenges. Current methodologies for quantifying the unpredictability metrics within these basins often involve computationally intensive processes due to the nonlinear operations that must be performed. Nevertheless, basins of attraction are well-suited for the application of machine learning techniques to predict the previously described metrics. Since machine learning algorithms infer nonlinear relationships from data, they are perfectly apt for the task.

The use of machine learning and deep learning techniques opens a wide field of applications for classifying basins of attraction. Indeed, machine learning has already been beneficial in solving problems within nonlinear dynamics. Among them, we can mention improving the prediction time of chaotic trajectories~\cite{fan2020long,pathak2018hybrid}, reconstructing the long-term dynamics of a system~\cite{lu2018attractor}, as well as classifying data with associated fractal properties~\cite{kirichenko2018machine, shi2018signal}. A first effort of automatic estimation of the fractal dimension of basins  has been recently published in~\cite{Valle2022}. 

The primary focus of this research is centered on utilizing Convolutional Neural Networks (CNNs) effectively to predict the selected unpredictability metrics of the basins. Our goal  is to leverage the capabilities of CNNs to streamline the computational challenges traditionally associated with conventional techniques. By doing so, we aim to enable swift and accurate forecasts of these pivotal metrics. This transition towards machine learning-driven methods not only simplifies the analytical process but also enriches our understanding of the predictability and stability of dynamical systems. This represents a significant advancement in this field, marking a noteworthy progression.

CNNs are a type of artificial neural networks designed specifically for image classification tasks. They were developed by LeCun et al.~\cite{726791} in 1998 as a class of deep feedforward artificial neural networks. They are composed of multiple layers whose characteristics are determined during the training process. Specialized layers detect features in the image, such as edges, shapes, and textures using multiple activation functions and weight matrices. The weights of the network are learned during the training stage, using labeled training data. This supervised learning approach is applied to the set of prelabeled basins acting as targets. Labels are simply the metrics of interest computed previously with a standard algorithm from the literature.

The rest of the paper is organized as follows. In Sec.~\ref{SystemsExplanation}, we provide insight about basins of attraction and their characterization metrics, followed by information about the computation of these metrics and the dynamical systems used. Section~\ref{pre-processing} presents the architectures of several CNNs used in our research, the image pre-processing, the image dataset distribution, and the hyperparameter settings for the training of the CNNs. Section~\ref{Results} presents a comparison of the performance of the proposed CNNs between themselves and reference methods for calculating the metrics, demonstrating the superior efficiency of our method.

\section{Complex dynamical systems and their characterization metrics}
\label{SystemsExplanation}

\begin{figure}[!h]
	\begin{subfigure}[]
		\centering
		\includegraphics[scale=0.22]{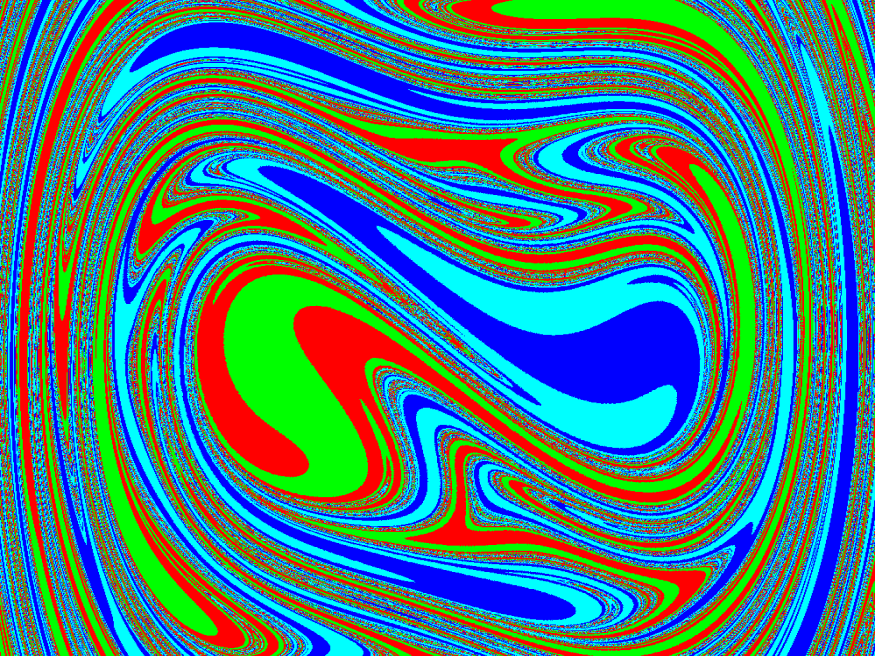}
		\label{Duffing_Basin}
	\end{subfigure}
	\begin{subfigure}[]
		\centering
		\includegraphics[scale=0.22]{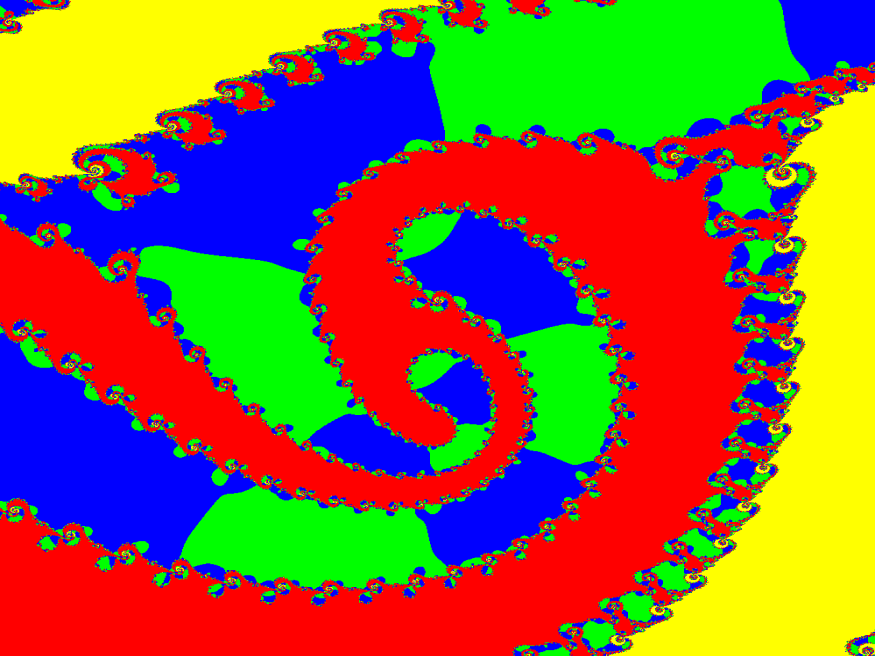}
		\label{Newton_Basin}
	\end{subfigure}
	\begin{subfigure}[]
		\centering
		\includegraphics[scale=0.22]{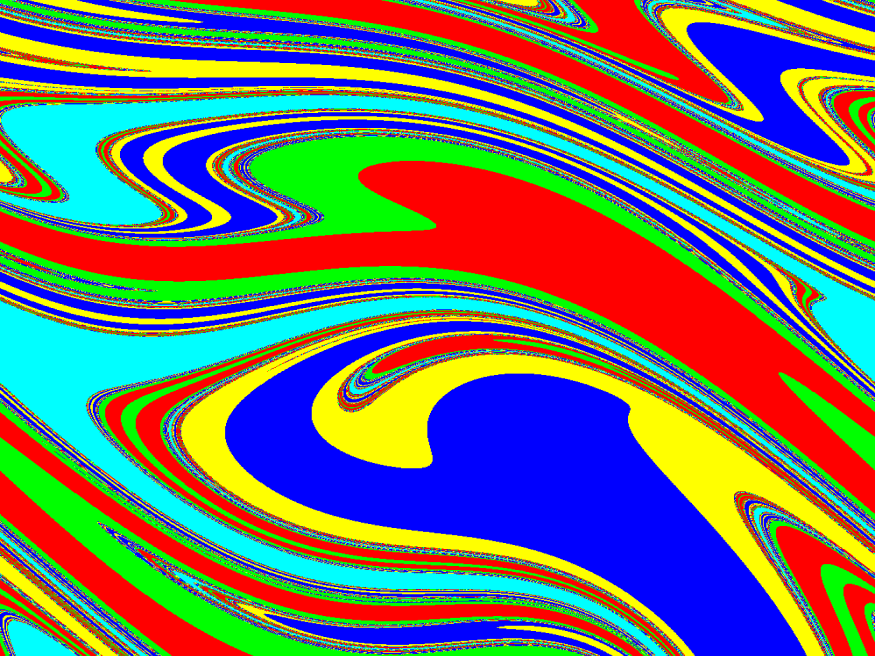}
		\label{PAF_Basin}
	\end{subfigure}
	\begin{subfigure}[]
		\centering
		\includegraphics[scale=0.22]{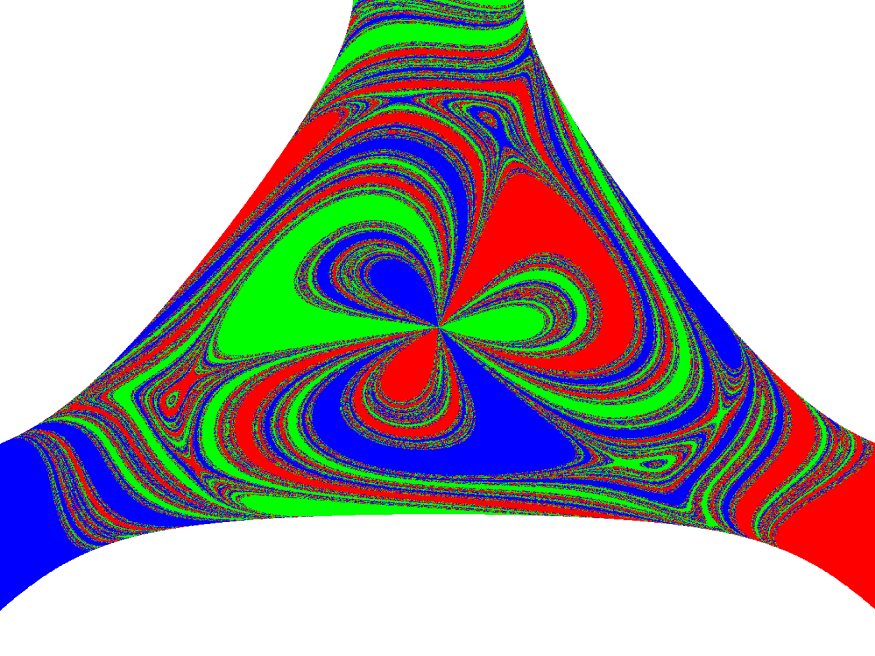}
		\label{HH_Basin}
	\end{subfigure}
	\begin{subfigure}[]
		\centering
		\includegraphics[scale=0.22]{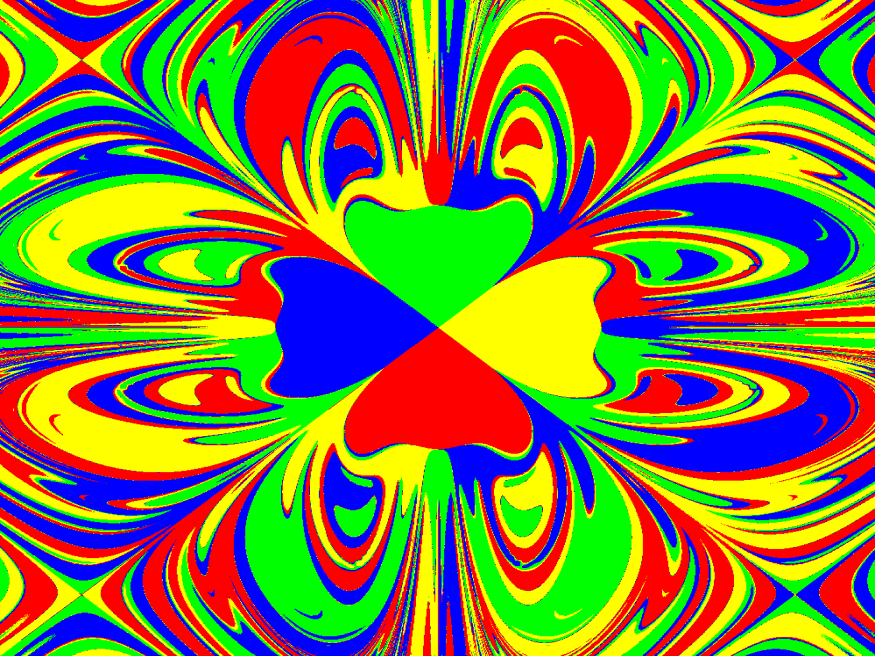}
		\label{MP_Basin}
	\end{subfigure}

\caption{\textbf{Examples of basins from the dynamical systems used in our work}. (a) This figure corresponds to a basin of attraction from the Duffing oscillator, (b) corresponds to a basin from the Newton fractal, (c) basins of attraction of the forced damped pendulum, (d) is an exit basin from the Hénon-Heiles Hamiltonian system, and finally (e) is an example of a basin of attraction of the magnetic pendulum with $4$ magnets. Further details about these dynamical systems can be found in Appendix~\ref{Appendix}.}
\label{BasinImages}
\end{figure}

The selected basins proceed from paradigmatic dynamical systems known for their ample variety of behaviors. We have computed basins of attraction from the Duffing oscillator, the forced damped pendulum, the Newton fractal, the magnetic pendulum, and finally the exit basins from the H\'{e}non-Heiles system (which in our work fulfill the same purpose as the basins of attraction). Examples of basins from these systems can be seen in Fig.~\ref{BasinImages}. Each basin consists of $333 \times 333$ trajectories for each choice of parameters of a system. Additional information about the computation performed here can be found in Appendix~\ref{Appendix}.

We summarize the main features about the computation of the metrics selected for the study of the basins, and provide some details:
\begin{itemize}
\item The fractal dimension of basins of attraction is a positive real number  FDim $\in [D-1 , D]$ where $D$ is the dimension of the phase space. For $D=2$, fractal basins are characterized by FDim $< 2$, meaning that the structures have a dimension somewhere between the line and the plane. To compute this number, we have implemented a Monte Carlo adaptation of the box-counting method proposed by McDonald et al.~\cite{mcdonald1985fractal}. Briefly, the method consists of detecting if a box of side length $\varepsilon$ located at random contains a piece of the boundary. If it does, the box is labeled as uncertain. The ratio of uncertain boxes over the total number of boxes $f(\varepsilon)$ depends on the box size. The fractal dimension is computed with a linear fit of the curve $\log f(\varepsilon)$ against $\log \varepsilon$. In our implementation, the fractal dimension was computed by randomly throwing boxes of side length ranging from $3$ to $33$ pixels on the basin. The box length size was increased by $3$ pixels every $350000$ boxes

\item The basin entropy is associated to the degree of uncertainty of the basins represented by a positive number which satisfies Sb $ \in [0 , log(N_a)]$ where $N_a$ refers to the number of attractors in the basins. For basins with Sb $= 0$ there is only one attractor, whereas for basins with Sb $= log(N_a)$ there are $N_a$ basins with intermingled boundaries~\cite{DazaClasifying2022}. To compute this number, we have implemented a Monte Carlo adaptation of the method proposed by Puy et al. in~\cite{Puy2021} where we measure the entropy of a box located at random in the basins, assuming that the trajectories inside that box are statistically independent. This measure is then repeated over a fixed number of random boxes, and the final basin entropy is the average entropy of all the boxes. In our implementation, the basin entropy was computed by randomly throwing $350000$ boxes of side length $15$ pixels.

\item The boundary basin entropy $Sbb$ is associated to the degree of uncertainty of the boundaries of the basins~\cite{DazaClasifying2022}. It is calculated using the same procedure as the basin entropy taking into account only the regions where a boundary is present. This metric provides a sufficient condition to assess easily that the boundaries are fractal, being $Sbb > log(2)$. This is a sufficient but not necessary condition, since there might be fractal boundaries with $Sbb < log(2)$. Nevertheless, this threshold can be very useful to assess quickly the fractality of some boundaries ~\cite{Daza2016}. 

\item Wada boundaries separate three or more basins at a time. From the dynamical point of view, the most interesting feature of Wada basins is the fact that an arbitrarily small perturbation of a system with initial conditions lying in a Wada boundary can drive the trajectory to any of the at least three possible attractors. It implies a special kind of unpredictability. In our work the presence of the Wada property was verified using the merging method proposed by Daza et al. ~\cite{Daza2018}. This method rely on the surprising fact that merging two of the Wada basins leaves unchanged the structure of the boundary. To assess the Wada property, it suffices to check that the boundary remains unaltered under the operation of merging any two basins. In our implementation of the merging method, the pixels of the merged boundaries and the original boundary are compared with some margin of error defined as the fattening parameter $r = 5$. 
\end{itemize}

We have computed each of these metrics for all our basins. The first three metrics were generated $10$ times for each basin since their calculations were done as a Monte Carlo method. By computing the same metric $10$ times for a basin, we have defined the true value of the metric as the mean value of those $10$ measurements in an effort to improve the robustness of measurements concerning their true mean value in subsequent iterations. The robustness and deviation of the measurements are discussed later in Sec.~\ref{Conclusions}. 

The number of basins for each system characterized with the conventional metrics are:
\begin{itemize}
\item Duffing oscillator: $84842$ basins,
\item Newton fractal: $10225$ basins,
\item Forced damped performed: $18926$ basins,
\item Hénon-Heiles: $1230$ basins, 
\item Magnetic pendulum: $14384$ basins. 
\end{itemize}
All these basins will be later used by the CNN to train and test for the accuracy of the predictions on each of the characterization metrics.

\section{Convolutional neural networks and their training}
\label{pre-processing}

The choice of a particular architecture for a classification task is not always obvious. The delicate balances and trade-offs between speed and accuracy depend on the network organization. The selected CNN architectures for the characterization were all breakthroughs in their respective fields, introducing novel techniques that improved the accuracy of image classification tasks. The architectures compared in our work are: AlexNet~\cite{NIPS2012_4824}, VGG16~\cite{simonyan2014very}, VGG19~\cite{simonyan2014very}, GoogLeNet~\cite{szegedy2015going}, and ResNet50~\cite{he2016deep}. Each architecture complexity and description is schematized in Table ~\ref{CNN_Architectures}.

To calculate the different characterization metrics, we have trained separate CNNs for each architecture mentioned above. Each metric has a dedicated trained CNN. We use the following terminology: FDim for the fractal dimension, Sb for the basin entropy, Sbb for the boundary basin entropy, and Wada for the presence of the Wada property.

The different architectures were implemented using the Python programming language libraries Tensorflow (2.7.0) and Keras (2.7.0). These libraries provide the building blocks necessary to develop the desired architectures, and all these were built following the same approach as explained by the authors of their corresponding papers~\cite{NIPS2012_4824, simonyan2014very, szegedy2015going, he2016deep}, with the exception of AlexNet and VGG architectures. In the case of AlexNet, the top dense layers were modified adding dropout~\cite{JMLR:v15:srivastava14a} to prevent overfitting and reducing the amount of neurons in half due to insufficient RAM memory. Whereas for the VGG architectures, the top dense layers were removed due to the poor results.
\begin{table}[!h]
	\begin{tabular}{c|c|c}
		Model & Trainable parameters &  Model description \\
		\hline AlexNet & $50.394.498$  & $5$ conv + $3$ fc layers \\
		VGG-16 & $14.824.386$ & $13$ conv + $1$ fc layers \\
		VGG-19 & $20.136.642$ & $16$ conv + $1$ $\mathrm{fc}$ layers \\
		GoogleNet & $17.675.382$ & $21$ conv + $1$ $\mathrm{fc}$ layers \\
		ResNet-50 & $23.630.722$ & $49$ conv + $1$ $\mathrm{fc}$ layers \\
	\hline
	\end{tabular}
\caption{\textbf{Comparison between the classical CNN architectures used in this article.} "conv" and "fc" in the model description denote convolutional and fully-connected layers. Each architecture follows the methodologies outlined in their respective papers \cite{NIPS2012_4824, simonyan2014very, szegedy2015going, he2016deep}, except for variations in the fully connected layers: AlexNet uses 2048 neuron layers, while VGG excludes these dense layers. Notably, GoogleNet has multiple outputs at different depths. Our research found that the deepest branch performs best across all metrics, hence subsequent results are based on this branch.}
\label{CNN_Architectures}
\end{table}

Regarding the image preprocessing, after the transformation of the basins to images, we started with matrices sized $333$ $\times$ $333$ which values were the classified attractors corresponding to each initial condition. We then reshaped each basin to $333$ $\times$ $333$ $\times$ $1$ as an image with same width and height as the number of initial conditions on each axis of the basin. The extra dimension corresponds to a grayscale 8-bit color channel where each pixel intensity encodes the attractor.

Due to the number of basins used in the training process, loading the whole image set overflows the RAM. Hence, we have developed a custom generator that has allowed us to randomly select batches of $16$ images, eventually granting the CNN to train with all the images in the set. Additionally, when forming a batch for the training, the images are subjected to two possible data augmentation techniques. They correspond to a horizontal and a vertical flip, each of them occurring with a probability of 50$\%$. So even by randomly sampling the training set, in case that the same image is repeated for the training, we still have a 50$\%$ chance of getting flipped horizontally, and a 50$\%$ of getting flipped vertically, or a 25$\%$ chance getting flipped both ways.

A careful design of the training, validation, and test sets is essential for the quality of the classification. The basins from the Duffing oscillator and the Newton fractal compose the training image set. The basins from the forced damped pendulum and the Hénon-Heiles Hamiltonian system are used as the validation image set, which leaves the basins from the magnetic pendulum to compose the test image set. 

\begin{figure}[!h]
\includegraphics[scale=0.35]{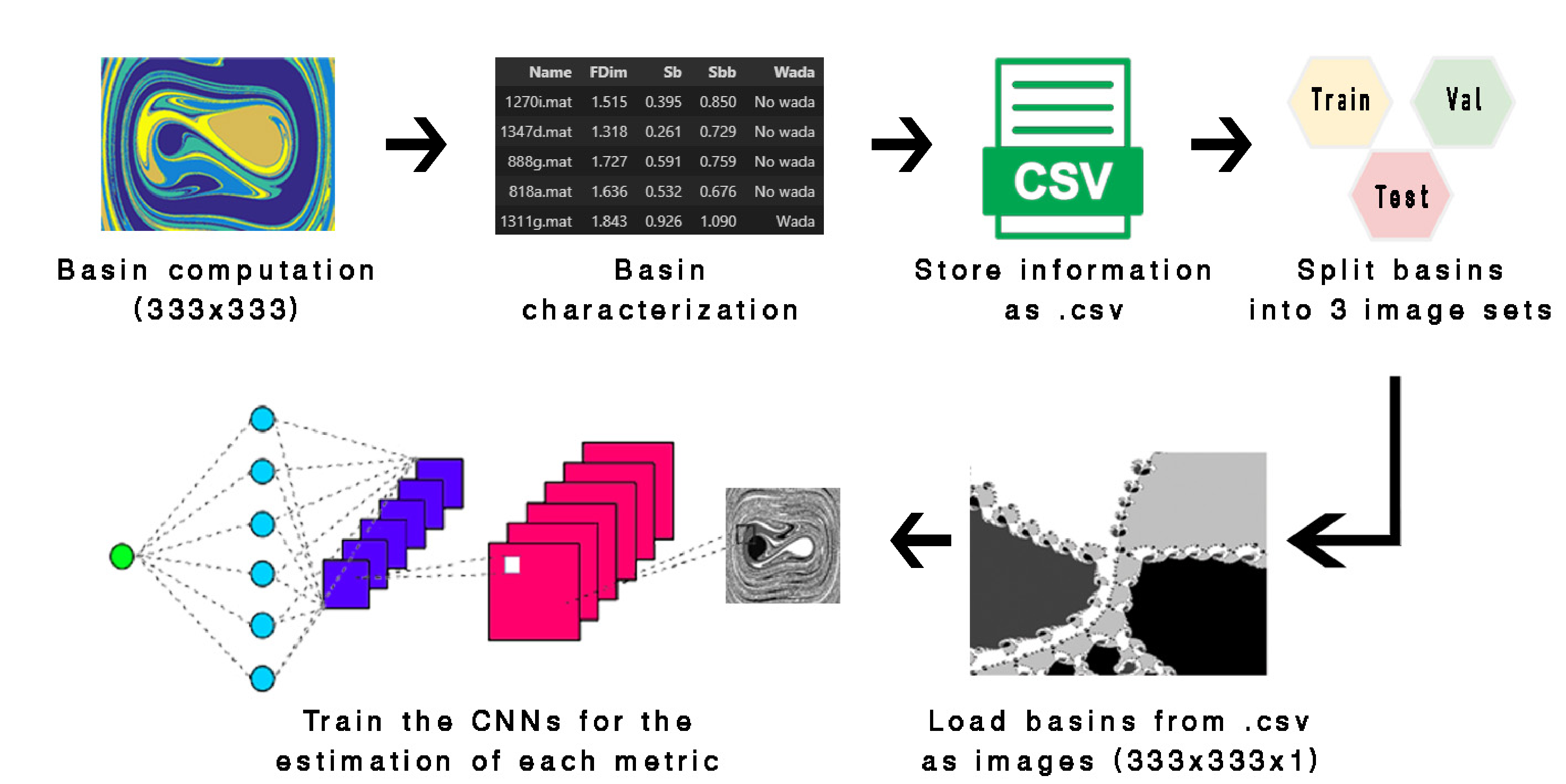}
\caption{\textbf{Schematic representation of the workflow followed in this work}. First, we compute basins from various dynamical systems and save their metrics and file paths in a dataframe, which is then stored as a .csv file. This file is utilized later by the CNN to access the basin information. Three distinct sets of images are taken into account: training, validation, and testing, with each computed basin being allocated to one of these image sets. During training, batches of 16 random basins are loaded from the .csv file, reshaped into grayscale images, and fed into the CNN. Four identical CNNs have been trained to accurately predict the desired metrics after training.}
\label{Pipeline}
\end{figure}

The distribution of basins has been tailored to balance the representation of each system across the training, validation, and test sets. The process of inputting an image, starting from its computation to its integration into the CNN, is schematized in Fig.~\ref{Pipeline}.

\onecolumngrid\
\twocolumngrid\
With this distribution, the training set for the neural network amounts to $73.35 \%$ of the total images with $95067$ images. The validation set was kept for improving the results through the validation step and amounts $15.55 \%$ of the total $20156$ images. And finally, the test set quantifies the prediction accuracy of the network with the remaining $11.10 \%$ consisting of $14384$ basins.
It is noteworthy that, when assessing the performance of the CNNs with the test set, we employed $1000$ subsets, each containing $1000$ images randomly selected from the test image set. This procedure is undertaken to estimate the error associated with the predictive capabilities of the networks.

\begin{figure}[!h]
\includegraphics[scale=0.56]{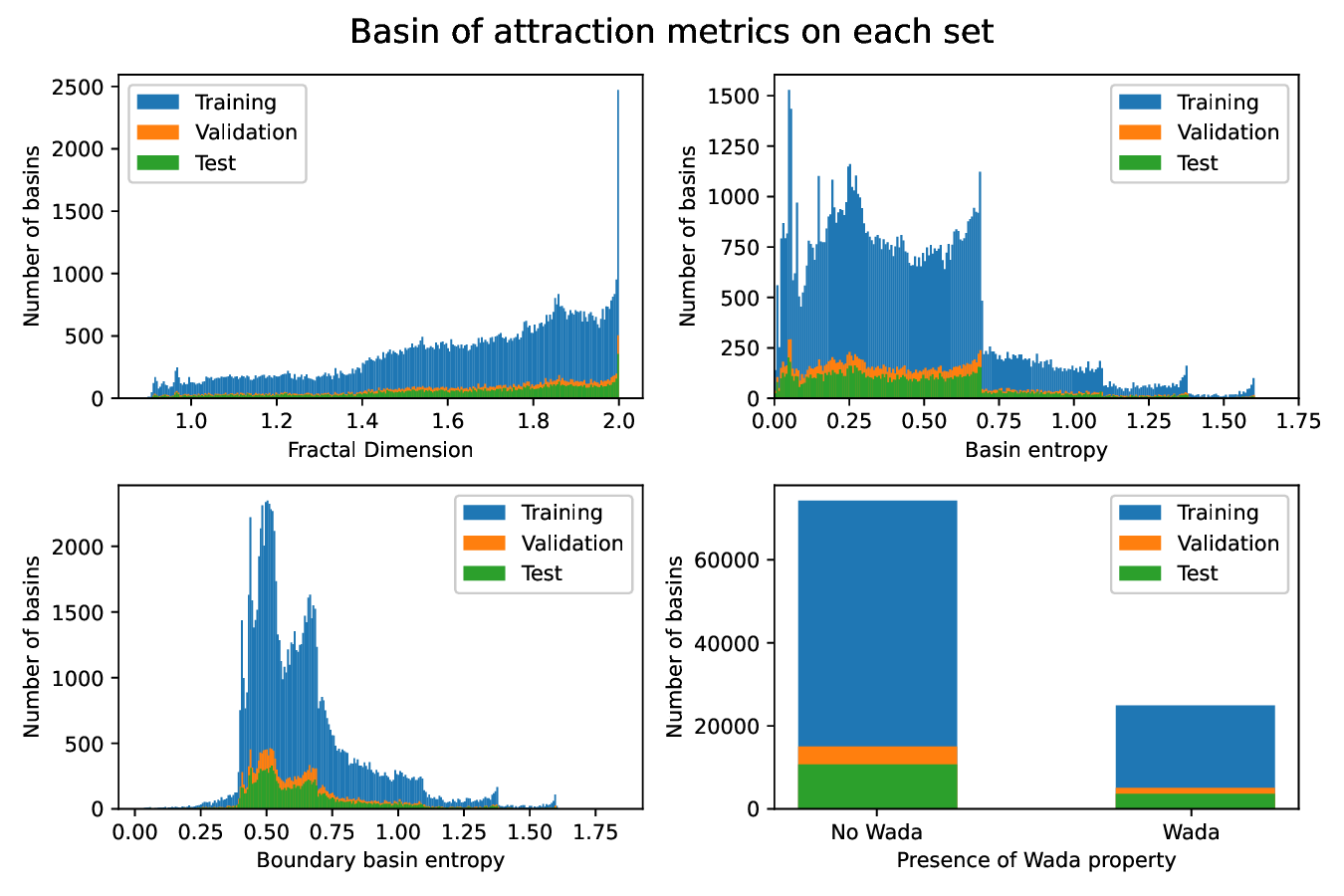}
\caption{\textbf{Distribution of values of each metric on all image sets.} The blue histogram denotes the training set, the orange denotes the validation image set, and the green denotes the test image set. It can be seen that the histograms do not follow a uniform distribution, which indirectly introduces a bias during training. However, by having the three sets with a similar distribution, the bias is minimized in our results.}
\label{all_set_metrics}
\end{figure}

The choice of parameters and systems influences the distribution of the metrics values. However, as can be observed in Fig.~\ref{all_set_metrics}, in spite of these uneven metrics, the distributions of the three sets are quite similar. Therefore the neural network is able to train accordingly and shows precise results.

With regards to the training of the different neural networks, an Nvidia Quadro RTX 5000 GPU was used. The weights and biases of the networks were randomly initialized and adjusted during the training through the optimizer provided by the tensorflow library \textit{Adam}~\cite{kingma2014adam} (Adaptive Moment Estimation Method). This optimizer was used with the default parameters: \textit{learning rate} $ = 0.001 $, $ \beta_1 = 0.9 $, $ \beta_2 = 0.999 $, $ \epsilon = 10 ^ {-8} $, $\text{decay} = 0$, $ amsgrad $ = true.

We trained each CNN for $100$ epochs while setting a batch size of $16$ images. Beyond this number of epochs we have noticed that the CNNs started to overfit. During this process we have measured the loss function of the validation image set as the mean square error (mse), when estimating the fractal dimension, the basin entropy, and the boundary basin entropy. The training step is designed to minimize the loss function. The mathematical expression is given by:
\begin{equation} 
\text{mse} = \frac{1}{N}\sum_{t=1}^{N}\left(\hat{x}_{t}-x_{t}\right)^{2},
\label{mse_formula}
\end{equation}
where $N$ is the total number of observations, $\hat{x_t}$ is the predicted value, and $x_t$ the expected value.

When checking the Wada property, the mean square error cannot be used as a loss function, since basins are labeled as either Wada or not Wada. A more useful loss function for this case is the categorical cross-entropy (cce). This loss function measures the difference between the predicted probability distribution and the true probability distribution of the target class, given by
\begin{equation}
\text{cce} = -\frac{1}{N} \sum_{t=1}^N \sum_{c=1}^C y_{tc} \log \left(\hat{y}_{tc}\right).
\label{categorical_crossentropy}
\end{equation}

From this expression, $N$ is the total number of observations, $C$ the total number of classes on the set (Wada or not Wada), $y_{tc}$ is the observation belonging to class $c$, and finally $\hat{y}_{tc}$ is the predicted observation $t$ belonging to class $c$.

Once the training is finished, the final set of weights and biases that performed best for the validation set were chosen as the final result of the training. All the code for the training, and a GUI to use the trained CNNs can be found in: \href{https://github.com/RedLynx96/Deep-Learning-based-Analysis-of-Basins-of-Attraction}{Github repository: https://github.com/RedLynx96/Deep-Learning-based-Analysis-of-Basins-of-Attraction}.

\onecolumngrid\
\twocolumngrid\
\clearpage

\section{Results and Performance}
\label{Results}

The most precise results of the measured metrics in each different architecture are displayed in Table \ref{mse_table}. They show the mean square error for the fractal dimension (FDim), the basin entropy (Sb), and the boundary basin entropy (Sbb), while for the Wada property we show the categorical cross-entropy. All these measurements are made on images of the validation set which, as specified in Sec.~\ref{pre-processing}, consist of basins from the forced damped pendulum and the Hénon-Heiles Hamiltonian system.

\begin{table}[!h]
\begin{tabular}{c|c|c|c|c}
Architecture& FDim $(10^{-4})$ & Sb $(10^{-4})$& Sbb $(10^{-4})$& Wada $(10^{-2})$\\ \hline
AlexNet & 1.756 & 4.258 & 2.103 & 0.719 \\
VGG16 & 3.668 & 0.863 & 3.980 & 4.214 \\
VGG19 & 3.692 & 1.408 & 3.406 & 0.709 \\ 
GoogLeNet & 1.015 & 1.024 & 5.924 & 2.305\\
\textbf{ResNet50} & $\mathbf{0.464}$ & $\mathbf{0.240}$ & $\mathbf{0.463}$ & $\mathbf{0.284}$\\
\hline
\end{tabular}
\caption{\textbf{Performance of each architecture for estimating the desired metrics on the validation dataset.} The metric shown in the table for the fractal dimension (FDim), the basin entropy (Sb), and the boundary basin entropy (Sbb) is the mean square error value at the end of the training for the validation set, whereas for the Wada property we the categorical cross-entropy. It can be seen from the table that the ResNet50 architecture is the most accurate one for all the metrics, as it has the least associated error.}
\label{mse_table}
\end{table}

As shown in the table, the ResNet50 architecture exhibits the lowest error. It can also be inferred in the cases of the estimation of the mean squared error, that the error of the neural networks increases with deeper network architecture except in the cases of GoogLeNet, and ResNet50. In the case of GoogLeNet, note that this architecture has three outputs and therefore each basin is predicted three times over a metric. Overall, the best performing branch of the architecture is the deepest one, and therefore all the results shown in this table belong to the output of the deepest branch. In Ref.~\cite{he2016deep}, the authors argued how deep convolutional neural networks make good predictions up to a certain point, where deeper networks involve more error in the predictions. This is the key point of ResNet, where a convolutional neural network with residual terms avoids this problem.

\begin{figure}[!h]
\includegraphics[scale=0.43]{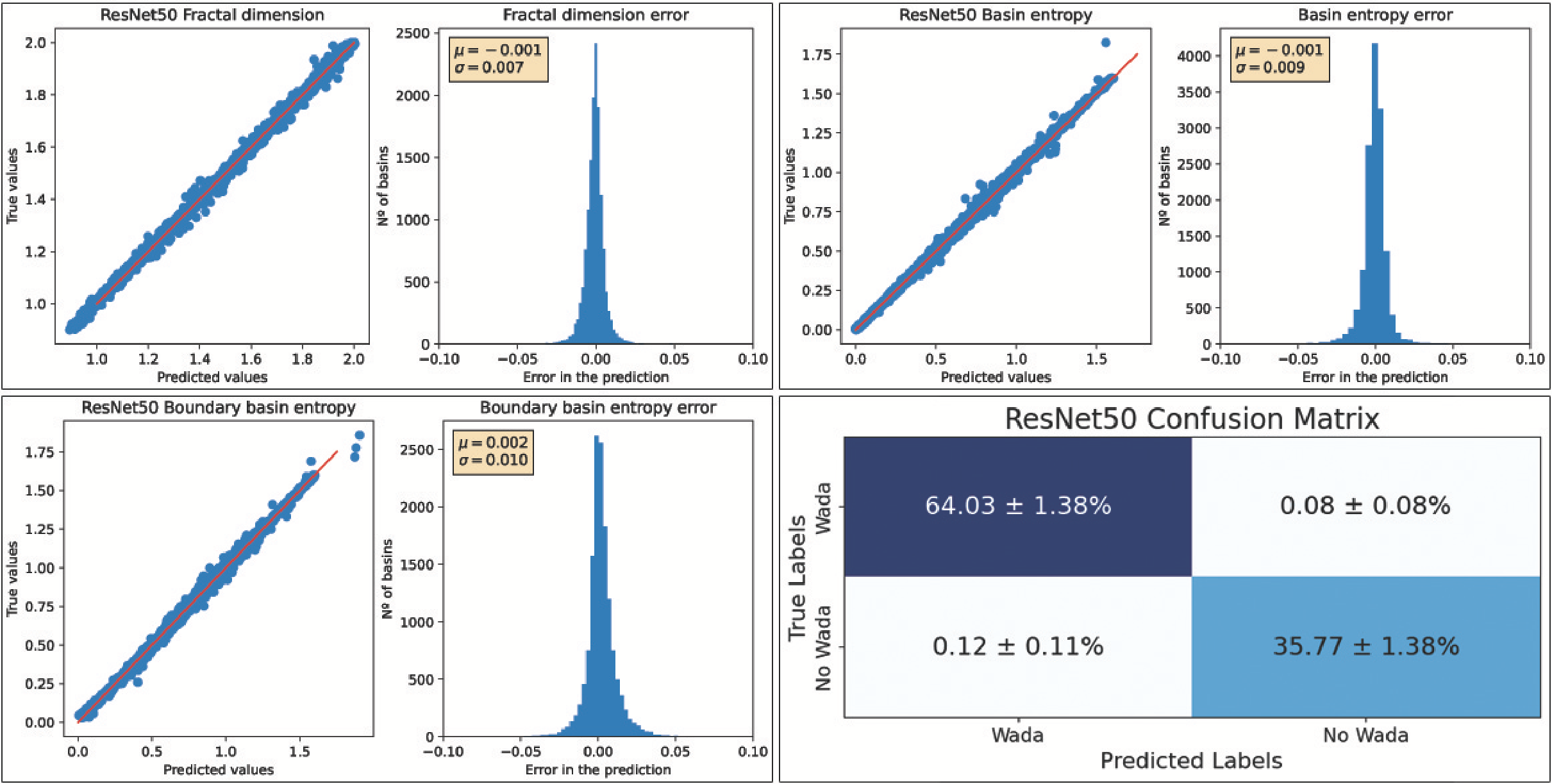}
\caption{\textbf{Performance of the ResNet50 architecture at predicting the $\mathbf{14384}$ basins of attraction from the test set}. The linear regressions show the comparison between the true values and the predicted values for each metric studied in our work. Each regression is followed in its right side with a histogram where the difference of the true and predicted values is shown. Such histogram approximates to a Gaussian distribution with mean $\mu$ and standard deviation $\sigma$. Since the estimation of the Wada property is a classification task, a more practical representation is the confusion matrix between the predictions and the true values. Such confusion matrix is shown in the bottom right of the image, showing the mean value of the accuracy for the prediction of $1000$ subsets of the test set, each comprising $1000$ random basins, yielding an accuracy of $99.8 \pm 2.76 \%$. proving the efficiency of the ResNet50 architecture.}

\label{ResNet50_all}
\end{figure}

Regarding the performance of our CNNs on the testing set, we used $14384$ images from the basins of attraction of the magnetic pendulum. We ran this image set through all our CNNs to assess their performance. To achieve this, we have computed all the metrics of each basin using all of our CNNs and compared the predicted values with the true values established by the reference methods. We have obtained the error in the prediction of each metric for a given basin by subtracting the predicted and true values, and then calculated the average error and its standard deviation for the $14384$ images. In Fig.~\ref{ResNet50_all}, we show the predictions of our yet to prove best performing architecture: the ResNet50. Here we compare the

\onecolumngrid\
\twocolumngrid\

\noindent predicted and true values of all the metrics in the test image basins, where the ideal case is to obtain a straight line which would indicate that the predicted and true values are the same. These linear regressions are followed by a histogram showing the distribution error of each prediction. With regard to the Wada property, a more useful approach is shown in the bottom right of Fig.~\ref{ResNet50_all}. The confusion matrix reflects how each basin of attraction has been classified and how it should have been classified, being the elements of the diagonal correctly classified and those outside wrongly classified. As mentioned in Sec.~\ref{pre-processing} we conducted $1000$ measurements of this metric on sets of $1000$ random images from the test set, therefore obtaining an estimation of the quality of the prediction for several samples of basins from the test set.

Finally, speed and performance tests for each CNN are shown in Table~\ref{test_performance_table}, where we show every architecture tested alongside the mean value error in the predictions $(\mu)$, the standard deviation of the errors $(\sigma)$, and the time $(t)$ required to characterize each basin from the testing set for the first three metrics. Whereas for the Wada property we show the mean accuracy of the $1000$ subsets of test basins analyzed $(\text{Acc})$ alongside the standard deviation of this value  $(\sigma)$. These results are also compared to the reference methods used to label our ground truths. As discussed in Sec.~\ref{SystemsExplanation}, the first three metrics are computed $10$ times for a single basin from where the mean value and standard deviation are calculated. The computation of the reference value of the Wada property do not convey error since they are computed with Monte Carlo methods.

The results in Table~\ref{test_performance_table} show that the architecture with the fastest computation time is AlexNet, which makes sense since it is the simplest architecture among those compared. The next fastest architecture is ResNet50, and even though it is not the fastest architecture, it is the one with the minimal error in its predictions. Therefore, we consider it to be the optimal architecture for performing this task, as the relationship between computation time and precision is the most advantageous among all the proposed architectures.

Although the ResNet50 has a slightly higher error compared to the reference methods, it provides a significant improvement in computation time. The ResNet50 characterizes basins much faster, taking only 7 minutes and 25 seconds, compared to a total time of 42 hours 49 minutes 38 seconds using the reference methods. This makes the ResNet50 approximately $\mathbf{342}$ times faster for the calculation of the $14384$ images from basins of attraction of the magnetic pendulum. Among the implemented solutions, this is the optimal architecture for a massive characterization of basins of attraction, such as exploring the evolution of a dynamical system.

One way to improve the accuracy and robustness of these results would be to improve the training conditions and labeling of the basins. As shown in Fig.~\ref{all_set_metrics}, although the histograms of the training, validation, and test sets have the same shape in each of the metrics, the distribution should be uniform to produce optimal results in the output of the network. To minimize the error in the predictions, it would be necessary to compute more basins from different dynamical systems so that the proportion remains balanced. However, achieving uniformity in one metric does not necessarily lead to uniformity in all of them, and to achieve this a greater number of basins and dynamical systems would be necessary.

The fractal dimension, the basin entropy, and the boundary basin entropy are metrics calculated via the Monte Carlo method. Hence, part of the error associated with these labels is due to the random sampling conditions. One way to minimize this error is achieved by measuring the same magnitude several times and taking the mean as the true value. Finally, regarding the Wada property, as we have explained before, we believe that the main source of error in the estimation of the metric comes from our ground truth. So a way to improve the utility of our work is labeling these basins with a different method such as the Nusse-Yorke~\cite{Nusse1996} or the saddle-straddle~\cite{Wagemakers2020}, which despite being more computationally expensive classifies basins in a more robust way.

\onecolumngrid\
\begin{table*}[!h]
\centering
\resizebox{1.00\textwidth}{!}{%
\begin{NiceTabular}{@{}c|ccc|ccc|ccc|cccc@{}}
& \Block{1-3}{FDim} & & & \Block{1-3}{$\mathrm{Sb}$} & & & \Block{1-3}{$\mathrm{Sbb}$} & & & \Block{1-3}{Wada} & & &\\ 
 \hline & $\mu$ & $\sigma$ & $t(s)$ & $\mu$ & $\sigma$ & $t(s)$ & $\mu$ & $\sigma$ & $t(s)$ & $\operatorname{Acc}(\%)$ & $\sigma$ $(\%)$ & $t(s)$ \\
 AlexNet & 0.03 & 0.014 & 63 & 0.013 & 0.035 & 62 & 0.000 & 0.025 & 59 & 81.29 & 2.82 & 60 \\
 Vgg 16 & -0.001 & 0.021 & 140 & 0.008 & 0.015 & 137 & 0.005 & 0.035 & 137 & 97.10 & 2.79 & 139 \\
 $\operatorname{Vgg} 19$ & -0.002 & 0.020 & 169 & 0.001 & 0.022 & 166 & -0.009 & 0.033 & 165 & 98.68 & 2.73 & 166 \\
 GoogLeNet & 0.014 & 0.038 & 78 & -0.004 & 0.013 & 79 & 0.009 & 0.038 & 79 & 97.05 & 2.79 & 81 \\
 \bf{ResNet50} & \bf{-0.001} & \bf{0.007} & \bf{111} & \bf{-0.001} & \bf{0.009} & \bf{110} & \bf{0.002} & \bf{0.010} & \bf{112} & \bf{99.80} & \bf{2.76} & \bf{112} \\
Reference Algorithms & & $5.29 \cdot 10^{-4}$ & 68602 & & $7.59 \cdot 10^{-4}$ & 80269 & & $7.56 \cdot 10^{-4}$ & 80269 & & 5307 \\
\hline
\midrule 
                                                                                                                   
\end{NiceTabular}
}
    \caption{\textbf{Performance comparison among CNN architectures.} Each architecture predicts the fractal dimension (FDim), the basin entropy (Sb), the boundary basin entropy (Sbb), and the presence of the Wada property (Wada) for the test set of $14384$ images of basins of attraction from the magnetic pendulum. The predicted values are compared to the ground truth values established by the reference methods explained in Sec.~\ref{SystemsExplanation}, giving the mean ($\mu$) and standard deviation ($\sigma$) error for each architecture. The time ($t$) each architecture takes to predict all images is also included. It can be seen how ResNet50 has the lowest prediction error and is faster than all architectures except for AlexNet. However, ResNet50 still outperforms AlexNet in terms of precision, making it the best architecture for this task. Therefore, it can be concluded that ResNet50 is the best performing architecture for this task.}
    \label{test_performance_table}
\end{table*}
\twocolumngrid\
\clearpage

\section{Conclusions}
\label{Conclusions}

We have trained a variety of CNNs using basins of attraction of paradigmatic nonlinear dynamical systems. These CNNs proved to be capable of characterizing complex basins quickly and accurately. Additionally, we have compared the different architectures of CNNs in terms of speed, precision, and robustness between them and against the computational methods from the literature. In doing so, we found that the ResNet50 architecture was the most efficient for this task.

The primary outcome of our research, as shown by Table~\ref{test_performance_table}, indicates that the ResNet50 model performs best in analyzing the $14384$ images from the test set. Observing the results, we can see that although the ResNet50 error is slightly higher than the reference algorithms error for the examined metrics, it offers a significant reduction in  computation time.

From all the CNNs architectures used in this work, the optimal architecture for this problem is the ResNet50 due to its precision and speed. Even though all the methods proposed here need a previously computed basin of attraction, the heavy computational time required for a massive characterization of basins when exploring one or more dynamical systems, can be easily minimized by the deep learning tool. The trade-off between computing speed and loss of precision is worth it for this task.

\appendix
\section{Dynamical systems parameters}
\label{Appendix}
We discuss the algorithms and the choice of parameters used for the computation of the basins from the different dynamical systems that we had discussed. All the basins and metrics were computed using an Intel Core i7-10700 CPU 2.90 GHz. The basins of attraction for the Duffing oscillator and the forced damped pendulum were computed numerically on the Julia programming language using a standard RK45 integrator and the method proposed by Datseris and Wagemakers~\cite{Recursion}. On the other hand, basins for the Hénon-Heiles Hamiltonian system, the magnetic pendulum, and the Newton fractal were computed numerically on C++ using a standard 4th order Runge-Kutta method. These basins of attraction were initially computed with a resolution of $1000 \times 1000$ pixels. However, to improve the size of the image set, each basin was split into 9 basins of $333 \times 333$ pixels, and also the original basin was downsampled to that resolution, thus obtaining a total of $10$ basins of $333 \times 333$ pixels from an original $1000 \times 1000 $ basin. 

As for the Duffing image set, we have computed $84842$ images corresponding to basins of the Duffing oscillator defined by
\begin{equation}
\ddot{x}+0.15 \dot{x} - x+ x^{3}=\gamma \cos \omega t.
\end{equation}

We have considered the ranges $0.1 \leq \gamma \leq 0.5$ and $0.2 \leq \omega \leq 2.5$.  This choice of parameters in the differential equation describes the motion of a unit mass particle in a double-well potential of the form $V(x) = x^4/4 - x^2/2$ , where the particle also experiences a linear drag with the environment $(0.15)$ and a external periodic forcing ($\gamma \cos \omega t$).

Regarding the forced damped pendulum, $18926$ images of basins of attraction were obtained from different parameters of the dynamical system. The system is defined by the following equation
\begin{equation}
\ddot{\theta} + 0.2\dot{\theta} + \sin \theta = F \sin \omega t.
\end{equation}

Here, we have considered $0.5 \leq \omega \leq 2.0$ and $0.5 \leq F \leq 2.0$. This choice of parameters in the differential equation covers a variety of basins of attraction, such as simple non-fractal basins, riddled basins, or Wada basins.

For the Hénon-Heiles Hamiltonian system, we have computed $1230$ exit basins, which are analogous to the basins of attraction for open Hamiltonian systems~\cite{Nieto2019}. The Hamiltonian associated with the Hénon-Heiles system is defined by:
\begin{equation}
H=\frac{1}{2}\left(p_x^2+p_y^2\right)+\frac{1}{2}\left(x^2+y^2\right)+x^2 y-\frac{y^3}{3}.
\end{equation}

This system describes the motion associated with a star around a galactic center~\cite{Henon1964}, the parameters \mbox{$x$, $y$, $p_x$ , $p_y$} are associated with the positions and momenta of a particle in the system, and $H=E$ is a constant that represents the total energy of the system. For the computation of the exit basins, we have considered values of the energy $0.5 \leq E \leq 2.0$, where the initial launch position of the particle has been modified $(x,y)$ and the associated moments $(p_x, p_y)$ have been computed using the tangential shooting as in~\cite{Aguirre2001}.

Regarding the Newton's fractal, $14276$ basins of attraction have been obtained. The polynomial used to calculate the basins of attraction was a fifth degree polynomial
\begin{equation}
p(z) = a_0 + a_1z + a_2z^2 + a_3z^3 + a_4z^4 + a_5z^5,
\end{equation}
where $z$ is an initial condition in the complex plane and $a_n$ is a real random number that satisfies $0.0 \leq a_n \leq 1.0$. Each iteration of the initial condition $z$ is performed following the rule
\begin{equation}
z_{n+1}=z_n-b \frac{p\left(z_n\right)}{p^{\prime}\left(z_n\right)}.
\end{equation}
The basins of attraction have been computed for different initial conditions of $z$, given that $-2.5 \leq z \leq 2.5$, for both real and imaginary parameters. In addition, the parameter $b$ has also been considered as a complex number, such that $0.0 \leq b \leq 1.0$ for both its real and imaginary parts. Thus, the basins of attraction have been obtained by computing different fifth-degree polynomials with random values for both $a_n$ and $b$.

Finally, in regard to the magnetic pendulum $14384$ basins of attraction of this system have been used. The system consists of a pendulum oscillating above a plane with magnets that is also influenced by the air friction. The differential equations that govern the movement of the pendulum in the system are defined by
\begin{equation}
\begin{split}
&\ddot{x}  +b \dot{x}+0.2 x- \\ & \sum_{i=1}^n \frac{x_i-x}{\left(\left(x_i-x\right)^2+\left(y_i-y\right)^2+0.2^2\right)^{3 / 2}}=0 \\ \\
&\ddot{y} +b \dot{y}+0.2 y-  \\ & \sum_{i=1}^n \frac{y_i-y}{\left(\left(x_i-x\right)^2+\left(y_i-y\right)^2+0.2^2\right)^{3 / 2}}=0.
\end{split}
\end{equation}

These differential equations describe the position of the pendulum $(x,y)$ at a distance $0.2$ above the plane in which $n$ magnets are located, the position of these magnets is given by the coordinates $(x_i,y_i )$ as they are placed having all the same separation between them and at the same distance $a$ from the center of the plane, which attracts the pendulum due to gravity and being represented with the scalar $0.2$. We also assume that the pendulum has linear drag $b$, and that the magnetic force between a magnet and the pendulum is proportional to $1/r^3$.

To compute the basins of attraction of this system we have considered $-0.01 \leq b \leq 0.25$, and the distance from the center to the magnets $1.5 \leq a \leq 3.0$. We have computed basins with this variation of parameters for the case in which $2,3$ and $4$ magnets appear in the system.

\section*{ACKNOWLEDGMENTS}
This work has been financially supported by the Spanish State Research Agency (AEI) and the European Regional Development Fund (ERDF, EU) under Project No.~PID2019-105554GB-I00 (MCIN/AEI/10.13039/501100011033).


%
\end{document}